\documentclass[sigconf, screen,]{acmart}
\AtBeginDocument{%
  }

\usepackage{svg}

\usepackage{amssymb}

\usepackage[most]{tcolorbox}
\usepackage{enumitem}
\usepackage{multirow}
\usepackage{fontawesome5}
\usepackage{diagbox}
\usepackage{colortbl}
\usepackage{makecell}
\usepackage{xspace}
\usepackage{threeparttable}
\usepackage{url}
\usepackage{caption}
\usepackage{booktabs}
\usepackage{subfig}
\usepackage{graphicx}
\usepackage{xcolor} 
\usepackage{pifont}
\usepackage{balance}
\usepackage{amsmath}
\usepackage{booktabs} 
\usepackage{array} 
\usepackage{CJKutf8}
\usepackage{adjustbox} 
\usepackage{tabularx}
\usepackage[table]{xcolor}
\usepackage{multirow}
\usepackage{subcaption}
\usepackage{enumitem}

\usepackage{pifont,xcolor}

\newcommand{\cmark}{\textcolor{green}{\ding{51}}}   
\newcommand{\xmark}{\textcolor{red}{\ding{55}}}     
\newcommand{\cmid}{\textcolor{orange}{\ding{51}}}   

\setcopyright{acmlicensed}
\copyrightyear{2018}
\acmYear{2018}
\acmDOI{XXXXXXX.XXXXXXX}

\acmISBN{978-1-4503-XXXX-X/2018/06}

\renewcommand\footnotetextcopyrightpermission[1]{}

\begin{document}

\makeatletter
\renewcommand{\@fnsymbol}[1]{%
  \ensuremath{%
    \ifcase#1\or \dagger\or \ddagger\or \mathsection\or \mathparagraph\or 
    \|\or **\or \dagger\dagger\or \ddagger\ddagger\else\@ctrerr\fi
  }%
}
\makeatother

\title{AgentSense: LLMs Empower Generalizable and Explainable Web-Based Participatory Urban Sensing}


\author{
    Xusen Guo\textsuperscript{\rm 1}, 
    Mingxing Peng\textsuperscript{\rm 1}, 
    Xixuan Hao\textsuperscript{\rm 1}, 
    Xingchen Zou\textsuperscript{\rm 1}, \\
    Qiongyan Wang\textsuperscript{\rm 1},  
    Sijie Ruan\textsuperscript{\rm 2},
    Yuxuan Liang\textsuperscript{\rm 1} 
}
\authornote{Corresponding Author. Email: yuxliang@outlook.com. }
\affiliation{
  \institution{
  \textsuperscript{\rm 1}The Hong Kong University of Science and Technology (Guangzhou), Guangzhou, China \\
  }
  \institution{
  \textsuperscript{\rm 2}Beijing Institute of Technology, Beijing, China \\
  }
  \country{}
}
\email{{xguo796, mpeng060, xhao390, xzou428, qwang650}@connect.hkust-gz.edu.cn}
\email{sjruan@bit.edu.cn, yuxliang@outlook.com}

\renewcommand{\shortauthors}{Xusen Guo. et al.}

\begin{abstract}
Web-based participatory urban sensing has emerged as a vital approach for modern urban management by leveraging mobile individuals as distributed sensors. However, existing urban sensing systems struggle with limited generalization across diverse urban scenarios and poor interpretability in decision-making. In this work, we introduce \textbf{AgentSense}, a hybrid, training-free framework that integrates large language models (LLMs) into participatory urban sensing through a multi-agent evolution system. AgentSense initially employs classical planner to generate baseline solutions and then iteratively refines them to adapt sensing task assignments to dynamic urban conditions and heterogeneous worker preferences, while producing natural language explanations that enhance transparency and trust. Extensive experiments across two large-scale mobility datasets and seven types of dynamic disturbances demonstrate that AgentSense offers distinct advantages in adaptivity and explainability over traditional methods. Furthermore, compared to single-agent LLM baselines, our approach outperforms in both performance and robustness, while delivering more reasonable and transparent explanations. These results position AgentSense as a significant advancement towards deploying adaptive and explainable urban sensing systems on the web.
\end{abstract}



\keywords{Web Mining, Participatory Urban Sensing, LLM, Agentic AI}


\maketitle

\section{Introduction}

The Web has become a fundamental infrastructure for urban sensing, enabling large-scale data collection, coordination, and interaction between citizens, devices, and city systems. Web-based platforms and mobile applications allow individuals, such as commuters, ride-hailing drivers, and couriers, to act as distributed sensors, continuously contributing spatio-temporal data through their daily activities. This paradigm, often termed \textbf{Web-based Participatory Urban Sensing (WPUS)}~\cite{sakamura2015minaqn, airantzis2008participatory, ji2016urban}, transforms the Web from a medium of information sharing into an active channel for real-time urban intelligence, powering applications in traffic monitoring~\cite{du2014effective, wang2018city, wang2022spatiotemporal}, environmental sensing~\cite{yang2021urban, hasenfratz2012participatory, jiang2016citizen}, and infrastructure management~\cite{zhao2018urban, zhang2021streamcollab}. Compared with fixed sensor networks, which are costly and static, web-enabled crowdsensing offers scalability, adaptability, and operational efficiency, making it a cornerstone for responsive, data-driven smart city services~\cite{szabo2013framework, hviid2020participatory}.

\begin{figure}[!t]
        \centering
        \includegraphics[trim={5.5cm 2.75cm 5.5cm 2.5cm}, clip,width=\linewidth]{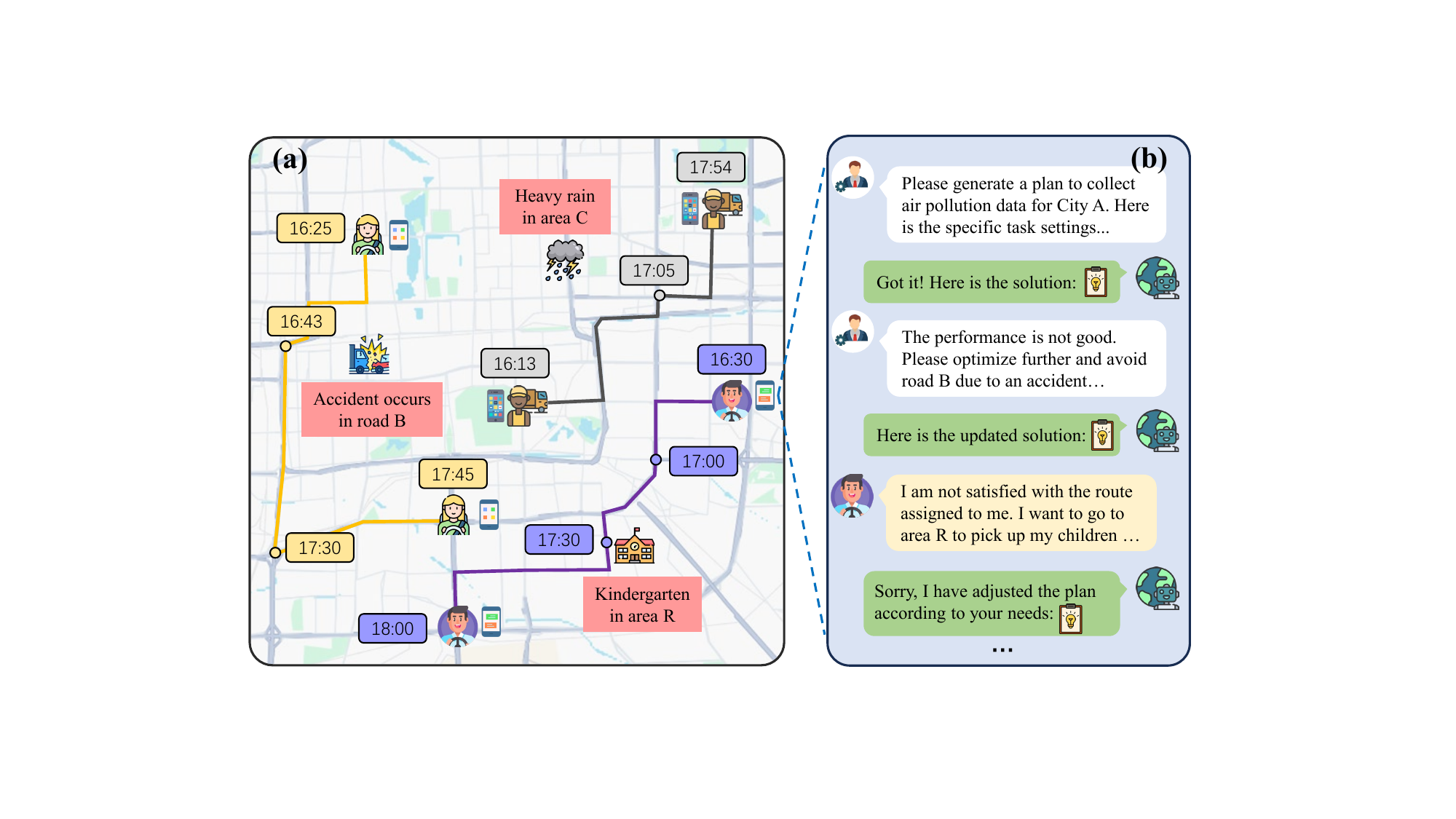}
        \vspace{-2em}
        \caption{(a): Web-based urban sensing under dynamic conditions. (b): AgentSense adaptively refines baseline solutions in response to evolving requirements. }
\label{fig:figure1}
\vspace{-2.0em}
\end{figure}

Early research on WPUS has made progress through optimization-based task allocation~\cite{ji2016urban, hasenfratz2012participatory, ji2023survey} and learning-driven approaches~\cite{wang2024urban}. The first class formulates the sensing assignment as a combinatorial problem under budget and data coverage constraints, while the latter introduces heuristics or reinforcement learning to improve scalability and adaptivity. These methods have proven effective in specific settings, but they remain largely limited to static assumptions and narrowly defined task environments. Despite progress, two critical challenges persist: 
i) \textbf{Context-Aware Generalization} is limited. Urban environments are inherently dynamic, shaped by real-time factors such as traffic, weather, and evolving worker availability (see (Figure~\ref{fig:figure1}(a)). Most systems cannot flexibly re-optimize task assignments under such disturbances~\cite{silva2013challenges, silva2014large}, and they often assume homogeneous worker behavior~\cite{ji2016urban, blunck2013heterogeneity}. Consequently, resources are used inefficiently, task distribution becomes inequitable, and models trained in one city fail to generalize to new contexts without costly retraining~\cite{ding2021crowdsourcing, wang2024urban}.
ii) \textbf{Transparent Interpretability} also remains unresolved. Current methods~\cite{xu2014gotcha, ji2016urban, wang2024urban} typically act as black boxes, offering no rationale for task reassignments or route changes. This lack of explanation undermines transparency and trust, particularly in safety-critical scenarios such as emergency response or public safety monitoring.
 
Recent advances in Large language models (LLMs)~\cite{ke2025survey, zeng2023evaluating, li2024survey} present new opportunities to overcome these challenges. For context-aware generalization, LLMs have demonstrated strong reasoning in complex, dynamic environments and robust zero-shot transfer across tasks and domains~\cite{kojima2022large,abbasiantaeb2024let}. These capabilities make them well-suited to integrate real-time urban disturbances with heterogeneous worker preferences, enabling adaptive and transferable task assignments. For transparent interpretability, LLMs naturally generate step-by-step reasoning and human-readable feedback, offering a pathway to explain why tasks are reassigned or routes adjusted, thereby improving trust in system decisions. However, directly applying LLMs to participatory urban sensing still remains infeasible, as \emph{the task assignment problem in WPUS is NP-hard, requiring combinatorial optimization beyond their generative capacity}.

To harness the complementary strengths of both worlds, we introduce \textbf{AgentSense}, a hybrid and training-free framework that integrates LLMs into WPUS through a \emph{multi-agent} loop. Our AgentSense \emph{combines the feasibility guarantees of classical planners with the generalizability and interpretability enabled by LLM-powered collaboration}. First, an initial solution is first generated by an optimization planner to satisfy fundamental constraints such as budget and coverage. This solution is then iteratively refined by three synergistic agents: a \textbf{Solver Agent} that proposes updates under dynamic disturbances, an \textbf{Eval Agent} that assesses solutions with quantitative metrics (such as coverage, budget compliance, and fairness) and structured feedback, and a \textbf{Memory Agent} that accumulates reusable meta-operations to accelerate convergence. As depicted in Figure~\ref{fig:figure1}(b), this design innovatively transforms WPUS into an interactive web-based platform that continuously adapts task assignments to evolving conditions while providing transparent explanations to participants and stakeholders. In summary, our contributions are fourfold:
\begin{itemize}[leftmargin=*,itemsep=0.1em,topsep=0em]
    \item \textbf{Dynamic and Personalized Task Assignment:} We propose AgentSense, a framework that adapts to real-time urban dynamics and heterogeneous participant preferences, enabling efficient, fair, and personalized task allocations.

    \item \textbf{Zero-Shot Generalization Across Tasks:} Leveraging the zero-shot capabilities of LLMs, AgentSense generalizes across diverse urban sensing tasks and environments without retraining, supporting rapid deployment in new scenarios.  

    \item \textbf{Transparency and Interpretability:} AgentSense enhances explainability by generating step-by-step refinements and natural language feedback, thereby improving transparency and fostering trust in urban management applications.  

    \item \textbf{Extensive Experimental Validation:} We conduct comprehensive experiments on two large-scale mobility datasets across seven types of disturbances, demonstrating that AgentSense consistently outperforms traditional methods in adaptivity and exceeds single-agent LLMs in both effectiveness and robustness. 
\end{itemize}

\section{Preliminary}

\subsection{Formulation}

\textbf{Definition 1 (Urban Sensing Problem)} The urban sensing problem involves recruiting a group of participants (or termed workers) and assigning tasks to maximize the coverage of data collected within a specified spatial-temporal region, subject to a given budget. Following work~\cite{ji2016urban}, the problem can be formulated as:
\begin{equation}
    \max{\mathcal{J}(D)}, \textit{~s.t.~} \Big \{D = \sum_{i \in P} D_{i}, \sum_{i \in P} r_i \leq B \Big \},
\end{equation}
where $P$ is the set of recruited participants, $D_{i}$ is the data collected by the sensing tasks assigned to participant $i$, $r_i$ is the rewards allocated to participant $i$, and $B$ is the total budget. The objective ${\mathcal{J}(D)}$ aims to maximize the overall data coverage, with each task defined as a location–time cell within a discretized city grid.

\vspace{0.1em}
\noindent\textbf{Definition 2 (Data Coverage)} We define the measurement of data coverage based on a hierarchical entropy-based objective function $\mathcal{J}(D)$ proposed in~\cite{ji2016urban}. The objective function is formulated as:
\begin{equation}
\mathcal{J}(D) = \alpha E(D) + (1-\alpha)\log_2{Q(D)},
\label{eq:coverage}
\end{equation}
where $E(D)$ represents the hierarchical entropy of the collected data $D$, which captures the balance of data at various spatial and temporal granularities, and $Q(D)$ is the total amount of collected data. The parameter $\alpha$ adjusts the relative importance between the balance of the data distribution and the total quantity of data. 

\vspace{0.1em}
\noindent\textbf{Definition 3 (Adaptive Urban Sensing Problem)} 
The adaptive urban sensing builds upon traditional approaches by incorporating real-time disturbances that may originate from environmental factors, participant-specific constraints, or system-level variations (e.g., road closures, adverse weather, or budget changes). The objective is to refine an initial urban sensing solution in response to these dynamic disturbances, ensuring feasibility and performance under changing conditions. Formally, the problem is defined as:
\begin{equation}\nonumber
    \max{\mathcal{J}(D(\delta), S_0)}, \textit{~s.t.~} \Big \{D(\delta)=\sum_{i \in P(\delta)} D_i(\delta), \sum_{i \in P(\delta)} r_i \leq B(\delta) \Big \},
\end{equation}
where $S_0$ represents the initial solution, $\delta$ denotes the disturbances or changes applied to the initial problem, and $D(\delta)$, $P(\delta)$, $B(\delta)$ are the total data collected, the participant candidates, and the total budget after the disturbance, respectively.

\begin{figure*}[!t]
        \centering
        \includegraphics[trim={1.3cm 2cm 2cm 1.5cm}, clip,width=0.98\linewidth]{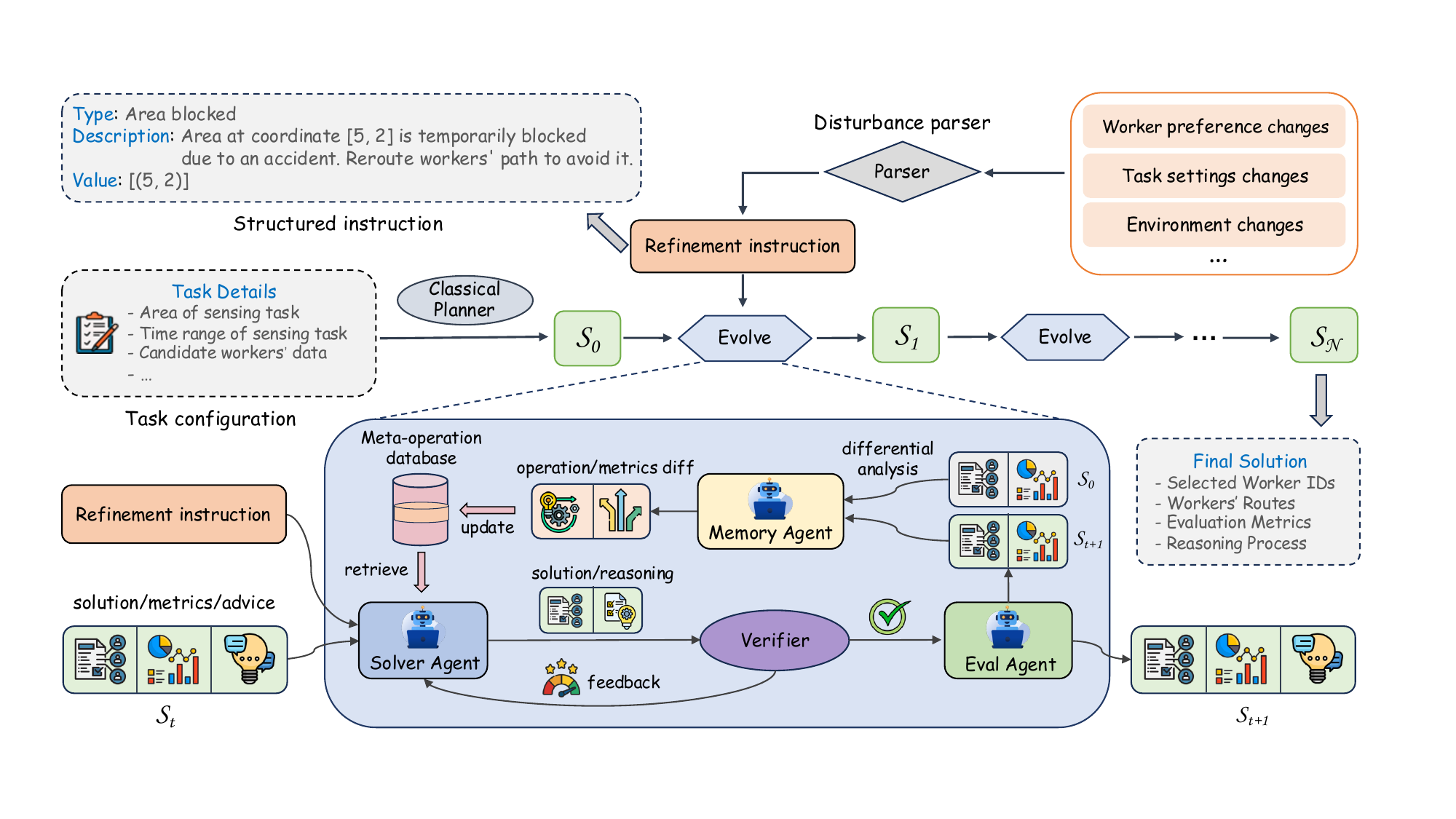}
         \vspace{-0.5em}
        \caption{The AgentSense framework for iterative refinement of urban sensing task solution.}
        \vspace{-0.5em}
\label{fig:framework}
\end{figure*}

\vspace{-0.5em}
\subsection{Related Work}

\subsubsection{Participatory and Mobility-Driven Urban Sensing} 


Participatory urban sensing has become a cornerstone of smart city research, enabling fine-grained monitoring of environmental conditions~\cite{hasenfratz2012participatory, jiang2016citizen}, transportation systems~\cite{wang2018city, wang2022spatiotemporal}, and urban infrastructure~\cite{zhao2018urban, zhang2021streamcollab}, etc. Recent studies have mainly focused on two paradigms: \textit{crowd-based human mobility sensing} and \textit{vehicle-based drive-by sensing}. The former treats humans as mobile sensors, where participants contribute GPS traces~\cite{ouyang2013if}, noise levels~\cite{zheng2014diagnosing}, or air quality data~\cite{liu2018third} during daily commutes. However, as noted in~\cite{ji2016urban}, human mobility is inherently skewed and uncertain, often leading to redundancy in popular areas while leaving peripheral regions under-sampled. To mitigate this imbalance, prior work has introduced entropy-driven objectives and graph-based task allocation methods under budget constraints~\cite{ji2016urban, wang2022privacy, wang2024urban}. In parallel, \textit{drive-by sensing} leverages vehicles such as taxis, buses, or drones equipped with sensors to provide broader and more systematic coverage~\cite{mathur2010parknet, agarwal2020modulo, ji2023survey}. While this approach offers improved reliability, its effectiveness is constrained by fixed mobility patterns, leading to route clustering and persistent blind spots in areas not served by the fleet~\cite{ghahramani2020urban, li2021two, xia2019identify}. Despite these strengths, both paradigms typically assume static mobility and lack mechanisms to adapt to dynamic real-world disturbances. In contrast, our AgentSense leverages LLM-driven evolution framework to continuously updates task plans, dynamically adapting to real-time changes.

\subsubsection{Large Language Models for Urban Computing.} 

LLMs have made significant progress in recent years, demonstrating strong capabilities in reasoning~\cite{ke2025survey, ferrag2025llm}, instruction following~\cite{zeng2023evaluating, qin2024infobench}, and multi-agent collaboration~\cite{li2024survey, hong2024metagpt, talebirad2023multi, han2024llm} across a wide range of domains~\cite{zhang2024towards, xi2025rise, aher2023using, guo2025automating, peng2025ld, argyle2023out}. Their strong capacity for natural language understanding, coupled with their alignment to human preferences, makes LLMs particularly suitable for urban applications that require translating human intent into system-level actions. Recent research has begun to explore this potential within urban contexts~\cite{zou2025deep, hou2025urban, liu2025urbanmind, li2025urban, li2024urbangpt}. In participatory planning, LLM agents can serve as both "planner" and "resident," synthesizing diverse stakeholder preferences into executable land-use strategies that outperform rule-based baselines and even human experts on service accessibility and ecological indicators~\cite{zhou2024large}. In the mobility domain, LLM-driven agents generate explainable, habit-aware daily trajectories that closely match real-world distributions and remain robust under disruptions like pandemics~\cite{wang2024large}. At the city scale, OpenCity~\cite{yan2024opencity} scales LLM agent simulations to tens of thousands of individuals through prompt optimization and system-level acceleration, enabling high-fidelity modeling of urban dynamics and counterfactual policy evaluation. Together, these advances highlight that urban contexts offers a particularly suitable ground for LLMs. Building on this perspective, we introduce AgentSense, a LLM-based multi-agent  framework for refining and adapting urban sensing plans under real-world disturbances.

\section{Methodology}

\subsection{Overview}

AgentSense is designed to adaptively handle urban sensing tasks in dynamic environments. As shown in Figure~\ref{fig:framework}, it follows a dynamic evolution workflow that begins with a classical planner generating an initial feasible schedule $\mathcal{S}_0$ over the road graph or discretized spatial-temporal grid. This plan ensures basic constraint satisfaction (e.g., time windows, total budget) and serves as a reference for subsequent refinements. To adapt to evolving conditions such as user preference updates or external disruptions (e.g., blocked roads, adverse weather), the system employs two key components: (i) a \emph{disturbance parser}, which converts unstructured dynamic signals into structured refinement instructions, and (ii) a \emph{multi-agent refinement loop}, where an LLM-powered multi-agent system iteratively updates baseline solution. This process yields a refinement sequence $\mathcal{S}_0 \to \mathcal{S}_1 \to \cdots \to \mathcal{S}_N$, progressively improving solutions while preserving spatial-temporal feasibility and budget compliance.

\subsection{Disturbance Parser}

Web-sourced signals from the environment or user feedback are inherently unstructured and semantically diverse. To enable automated adaption over these signals, the disturbance parser leverages a large language model to translate them into structured refinement instructions. Each instruction is formatted as a dictionary with standardized fields \texttt{type}, \texttt{description}, and \texttt{value}. These structured messages precisely encodes the scope and semantics of the required changes, enabling the downstream refinement process to operate over a unified, machine-interpretable interface.

\vspace{-1.0em}
\subsection{Solver Agent}

As the core component of the multi-agent refinement framework,  \emph{Solver Agent} is responsible for transforming a current solution $\mathcal{S}_t$ into an improved solution $\mathcal{S}_{t+1}$ in response to a structured refinement instruction $\delta$. Its primary objective is to incorporate disturbances while preserving feasibility and optimizing key criteria such as spatial-temporal coverage and budget efficiency.  To achieve this, the \emph{Solver Agent} operates through an \emph{iterative edit-and-verify} loop. At each step, it applies atomic and interpretable modifications (e.g., adding / removing workers, swapping task assignments), and verifies that the resulting plan remains valid. This fine-grained editing process ensures both transparency of reasoning and robustness.

While the \emph{Solver Agent} orchestrates the refinement, it is supported by two auxiliary sources of guidance: evaluation feedback provided by the \emph{Eval Agent} and historical meta-operations managed by the \emph{Memory Agent}. These additional information allow the solver to prioritize promising adjustments, avoid redundant modifications, and accelerate convergence. Over successive refinements, the \emph{Solver Agent} accumulates reusable heuristics, enabling more adaptive and efficient planning in dynamic urban sensing environments.

\subsection{Eval Agent}

\begin{figure}[!t]
        \centering
        \includegraphics[trim={1.5cm 7.5cm 18cm 5cm}, clip,width=\linewidth]{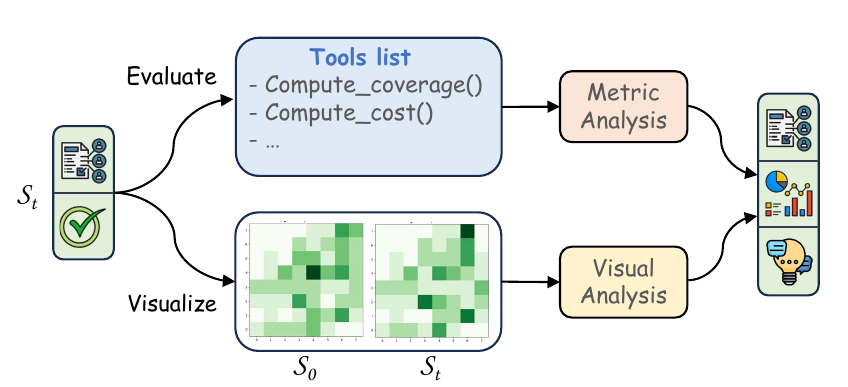}
        \vspace{-2em}
        \caption{Eval Agent.}
        \vspace{-1.4em}
\label{fig:eval_agent}
\end{figure}

The \emph{Eval Agent} functions as a quality assessor and feedback generator within each refinement iteration. As illustrated in Figure~\ref{fig:eval_agent}, given a candidate solution $\mathcal{S}_t$ proposed by the \emph{Solver Agent}, the \emph{Eval Agent} evaluates its performance across multiple quantitative criteria with external function tools, including total data coverage and budget efficiency. Beyond numerical metrics, it also generates spatial-temporal diagnostics by visualizing coverage heatmaps for both the baseline solution $\mathcal{S}_0$ and the refined solution $\mathcal{S}_t$. This visual comparison helps identify gaps, or redundant routing patterns that can not be evident from metrics alone. The agent then composes its findings into a concise natural language report, which includes a ranked list of actionable suggestions. Examples include "reroute worker 46 to pass through the low-coverage top-left region" or "replace worker 2 with worker 5 to reduce cost while maintaining coverage." This structured feedback is passed to the \emph{Solver Agent} to guide subsequent iterations, ensuring that the refinement process remains interpretable, goal-directed, and human-aligned.

\subsection{Memory Agent}

\begin{figure}[!t]
    \centering
    \includegraphics[trim={18cm 7cm 5cm 4cm}, clip,width=0.9\linewidth]{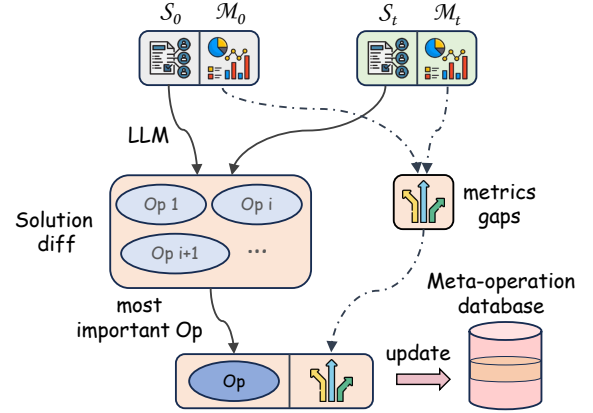}
    \vspace{-0.5em}
    \caption{Memory Agent.}
    \vspace{-1em}
\label{fig:memory_agent}
\end{figure}

The \emph{Memory Agent} enables experience-aware refinement by systematically recording and analyzing the solver’s historical optimization trajectories. Given a baseline–refined solution pair \((\mathcal{S}_0, \mathcal{S}_t)\) and their corresponding evaluation metrics \((\mathcal{M}_0, \mathcal{M}_t)\), the agent first computes the metric gap $\Delta \mathcal{M} = \mathcal{M}_t - \mathcal{M}_0$, which reflects the improvement or degradation induced by refinement. Next, the agent conducts a differential analysis between \(\mathcal{S}_0\) and \(\mathcal{S}_t\), extracting a set of interpretable atomic \emph{meta-operations}:
\[
\Delta \mathcal{S}_t = \{\, o_1, o_2, \dots, o_n \,\}, \quad 
o_i \in \mathcal{O},
\]
where \(\mathcal{O}\) is the operation space (e.g., \textit{reroute path segment $\varsigma$ of worker $u$}, \textit{add / remove worker $w$}, \textit{replace task $i$ with $j$}). Each operation \(o_i\) is then assigned an impact score
\[
\phi(o_i \mid \Delta \mathcal{M}, \mathcal{C}_t) \in \mathbb{R},
\]
estimated by the LLM given the observed metric gap \(\Delta \mathcal{M}\) and contextual metadata \(\mathcal{C}_t\) (e.g., budget, region, disturbance type). The operation with the highest score is identified as a high-impact meta-operation:
\[
o^* = \arg\max_{o_i \in \Delta \mathcal{S}_t} \phi(o_i).
\]

All high-impact operations, along with their context and associated \(\Delta \mathcal{M}\), are stored in a persistent database \(\mathcal{D}\). In subsequent refinement rounds, the \emph{Solver Agent} queries this database with its current candidate operations \(\Delta \mathcal{S}_{t'}\), retrieving the top-$k$ relevant past operations:
\[
\text{Retrieve}(\Delta \mathcal{S}_{t'}) = \operatorname{Top\text{-}k}_{o \in \mathcal{D}} \; \text{sim}(o, \Delta \mathcal{S}_{t'}),
\]
where $\text{sim}(\cdot,\cdot)$ denotes an embedding-based similarity between operations. By leveraging these retrieved experiences, the solver learns from prior refinements, reuses successful strategies, and avoids repetitive failures. This memory-guided mechanism accelerates convergence and improves overall solution quality.

\subsection{Comparison to Existing Arts}
\vspace{-0.5em}

\begin{table}[!h]
\centering
\renewcommand{\arraystretch}{1.0}
\caption{Comparison of different paradigms in WPUS with different perspectives. \textcolor{green}{\ding{51}} / \textcolor{orange}{\ding{51}} / \textcolor{red}{\ding{55}}  indicates strong / moderate / lack of support.}
\vspace{-0.5em}
\begin{tabular}{c|c|c|c|c}
\Xhline{1.0pt}
\textbf{Dimension} & Classical & \makecell{Learning\\-based} & \makecell{single \\ LLMs} & \textbf{AgentSense} \\
\Xhline{1.0pt}
Solution quality                 & \cmark & \cmark & \cmid & \cmark \\
Efficiency                  & \cmark & \xmark & \xmark & \cmid \\
Explainability              & \xmark & \xmark & \cmid & \cmark \\
zero-shot ability         & \xmark & \xmark & \cmark & \cmark \\
Scalability                 & \cmark & \cmid & \xmark & \cmid \\
\Xhline{1.0pt}
\end{tabular}
\label{tab:comparison}
\vspace{-0.5em}
\end{table}

To better position our framework within the broader landscape, we compare \textbf{AgentSense} with existing paradigms in WPUS. As summarized in Table~\ref{tab:comparison}, classical planners rely on explicit optimization rules, providing strong solution quality with high efficiency and scalability, yet their rule-based nature limits adaptability, resulting in poor explainability and a lack of zero-shot generalization. Learning-based approaches deliver high-quality solutions by leveraging data-driven modeling, but their dependence on training data reduces efficiency, limits scalability, and weakens interpretability. Single-agent LLMs, by contrast, offer explainability and zero-shot generalization through intrinsic reasoning ability, but often generate infeasible solutions and suffer from high computational costs in complex reasoning, which prevents scalability. By integrating classical planning with a multi-agent refinement loop, \textbf{AgentSense} combines structural reliability with LLM-driven adaptability, achieving consistently high solution quality while ensuring interpretability and generalization. Although its efficiency and scalability remain constrained by the computational burden of LLMs, AgentSense leverages baseline solutions to confine exploration within a feasible subspace, thereby achieving a balanced trade-off. 

\section{Experiments}

In this section, we evaluate the performance of \emph{AgentSense} across five key research questions:

\begin{itemize}[leftmargin=*]
\item \textbf{RQ1:} How significantly does AgentSense improve upon strong single-agent LLM baselines across diverse urban sensing tasks?
\item \textbf{RQ2:} Does better baseline planners lead to better performance?
\item \textbf{RQ3:}How effectively does AgentSense adapt to diverse real-world disturbances while maintaining solution quality?
\item \textbf{RQ4:} How do the \emph{Eval Agent} and the \emph{Memory Agent} contribute to overall efficiency and quality of the solution?
\item \textbf{RQ5:} How does AgentSense improve upon single-agent LLM baselines in terms of explanation?
\end{itemize}

To answer these questions, we conduct extensive experiments across diverse task settings and disturbances.

\subsection{Experimental Setup}

\subsubsection{Datasets}

We conduct experiments on two large-scale real-world mobility datasets, T-Drive~\cite{tdriver2010} and Grab-Posisi~\cite{grabposis2019}, which span diverse urban regions and mobility patterns. T-Drive contains GPS traces over ten thousands taxis in Beijing over a week in 2008, offering rich spatiotemporal signals for large-scale sensing. Grab-Posisi, released by Grab, records high-frequency positional data from ride-hailing drivers across cities like Jakarta, enabling fine-grained analysis in dense urban settings.

To evaluate our framework under different complexities, we define three configurations (\emph{Small}, \emph{Medium}, and \emph{Large}) for each dataset, as summarized in Table~\ref{tab:dataset_settings}. In the \emph{Small} T-Drive setting, for instance, the city is discretized into an $8 \times 8 \times 8$ spatial-temporal sensing grid (15-minute intervals over a 2-hour horizon), with each cell corresponds to a distinct urban sensing task. We sample 20 workers and allocate a total budget of 40, which specifies the maximum cumulative rewards distributed across workers. Following prior work~\cite{ji2016urban, wang2024urban}, workers are assumed to move at constant speed in free space with a normalized reward of 1.0 per time step. Across all settings, the sensing task time interval is fixed at 15 minutes for T-Drive and 5 minutes for Grab-Posisi. To ensure robustness, we random generate 20 task instances for each setting in each dataset.
\begin{table}[!h]
\footnotesize
    \centering
    \vspace{-1em}
    \renewcommand{\arraystretch}{1.0}
    \caption{Details of datasets.}
    \vspace{-1.0em}
    \begin{tabular}{p{1.05cm}|p{0.7cm}|c|c|c|c|c}
    \Xhline{1.0pt} 
    Dataset & Scale & Workers & Grid Size & \# Regions & \makecell{Total \\ Budget} & \makecell{Horizon \\ (min)}\\
    \Xhline{1.0pt} 
    \multirow{3}{*}{T-Drive} 
    & Small  & 20 & 8 $\times 8$  & 64 & 40  & 120\\
    & Medium & 40 & 16 $\times 16$ & 256 & 60  & 240\\
    & Large  & 60 & 32 $\times 32$ & 1024 & 100 & 360\\
    \hline
    \multirow{3}{*}{Grab-Posisi} 
    & Small  & 15 & 8 $\times$ 4  & 32 & 40  & 40\\
    & Medium & 30 & 16 $\times 8$ & 128 & 60  & 80\\
    & Large  & 45 & 32 $\times 16$ & 512 & 100 & 160\\
    \Xhline{1.0pt} 
    \end{tabular}
    \vspace{-1.8em}
\label{tab:dataset_settings}
\end{table}




\subsubsection{Evaluation Metrics}

To quantitatively assess the performance of AgentSense and baseline models, we adopt three evaluation metrics that capture both effectiveness and efficiency. \textbf{SR} (Success Rate) denotes the proportion of runs (out of 20 independent trials) where the refined solution achieves a valid improvement over the baseline within at most 10 iterations. \textbf{AIR} (Average Improvement Rate) measures the relative improvement in the main objective function (Equation~\ref{eq:coverage}, with $\alpha=0.5$ by default) compared to the baseline solution. \textbf{ANI} (Average Number of iterations) records the average number of refinement steps required to obtain a valid improvement, reflecting the convergence speed of the refinement process. Finally, \textbf{ACS} (Average Cost Savings) quantifies the reduction in budget usage, where negative values correspond to a cost increase.


\subsection{Overall Performance (RQ1)}

\begin{table*}[!t]
\centering
\renewcommand{\arraystretch}{1.0}
\caption{Comparison of single-agent LLMs and AgentSense on the T-Drive and Grab-Posisi problem sets across three task settings. SR denotes the success rate over 20 trials, and AIR represents the average improvement rate of the objective value.}
\vspace{-1em}
\begin{tabular}{c|c|cc|cc|cc|cc|cc|cc}
\Xhline{1.0pt}
\multirow{3}{*}{\makecell{\\ \textbf{Type}}} & \multirow{3}{*}{\makecell{\\ \textbf{Model}}} & \multicolumn{6}{c|}{\textbf{T-Drive}} & \multicolumn{6}{c}{\textbf{Grab-Posisi}} \\ \cline{3-14}
& & \multicolumn{2}{c|}{Small} & \multicolumn{2}{c|}{Medium} & \multicolumn{2}{c|}{Large} & \multicolumn{2}{c|}{Small} & \multicolumn{2}{c|}{Medium} & \multicolumn{2}{c}{Large} \\ \cline{3-14}
 & & \makecell{SR \\ (\%) $\uparrow$} & \makecell{AIR \\ (\%) $\uparrow$} & \makecell{SR \\ (\%) $\uparrow$} & \makecell{AIR \\ (\%) $\uparrow$} & \makecell{SR \\ (\%) $\uparrow$} & \makecell{AIR \\ (\%) $\uparrow$} & \makecell{SR \\ (\%) $\uparrow$} & \makecell{AIR \\ (\%) $\uparrow$} & \makecell{SR \\ (\%) $\uparrow$} & \makecell{AIR \\ (\%) $\uparrow$} & \makecell{SR \\ (\%) $\uparrow$} & \makecell{AIR \\ (\%) $\uparrow$} \\


\hline
\multicolumn{14}{c}{\cellcolor{gray!10}{\textit{Proprietary LLMs}}} \\
\Xhline{1.0pt} 
\multirow{5}{*}{SA}
& gpt-3.5-turbo-0125 & \underline{95.0} & 0.894 & \underline{90.0} & 0.202 & \underline{70.0} & -0.541 & \underline{95.0} & 0.934 & 85.0 & 0.185 & \underline{75.0} & -0.453 \\
& gpt-4.1-2025-04-14 & 70.0 & \underline{1.352} & 55.0 & 0.982 & 30.0 & 0.091 & 75.0 & \underline{1.503} & 60.0 & 0.880 & 50.0 & 0.105 \\
& gemini-2.0-flash   & \textbf{100.0} & 0.904 & 85.0 & 0.320 & 65.0 & 0.084 & 95.0 & 0.864 & \underline{90.0} & 0.240 & 65.0 & 0.232 \\
& claude-3-7-sonnet  & 65.0 & 1.165 & 55.0 & \underline{1.059} & 45.0 & \underline{0.722} & 70.0 & 1.244 & 55.0 & \underline{1.123} & 50.0 & \underline{0.504} \\
& claude-3-5-sonnet  & 65.0 & 0.620 & 50.0 & 0.455 & 25.0 & 0.246 & 65.0 & 0.598 & 55.0 & 0.401 & 35.0 & 0.277 \\
\hline
\multicolumn{14}{c}{\cellcolor{gray!10}{\textit{Open-Source LLMs}}} \\
\Xhline{1.0pt} 
\multirow{3}{*}{SA}
& deepseek-r1           & \textbf{100.0} & 0.520 & 85.0 & -0.183 & \underline{70.0} & -0.677 & \textbf{100.0} & 0.497 & \underline{90.0} & 0.092 & \underline{75.0} & -0.744 \\
& qwen3-235b-a22b       & \textbf{100.0} & 0.433 & 70.0 & 0.370 & 55.0 & 0.012 & \underline{95.0} & 0.465 & 70.0 & 0.401 & 50.0 & -0.043 \\
& Llama-3-70B-instruct  & 70.0 & 1.208 & 85.0 & 0.314 & 65.0 & 0.087 & 75.0 & 0.993 & 80.0 & 0.287 & 55.0 & 0.094 \\
\Xhline{1.0pt} 
MA & 
\textbf{AgentSense} & \textbf{100.0} & \textbf{1.545} & \textbf{100.0} & \textbf{1.162} & \textbf{95.0} & \textbf{0.983} & \textbf{100.0} & \textbf{1.622} & \textbf{100.0} & \textbf{1.248} & \textbf{100.0} & \textbf{0.648} \\
\Xhline{1.0pt}
\end{tabular}
\label{tab:baselines}
\end{table*}

To underscore the advantages of AgentSense over single-agent LLM baselines, Table~\ref{tab:baselines} reports a comparative evaluation on two datasets across all three task settings. As expected, all models exhibit notable performance drops from the \emph{Small} to the \emph{Large} setting, validating the effectiveness of our task hierarchy in reflecting increasing complexity. Furthermore, performance trends remain consistent across both datasets, indicating strong generalization of LLM-based approaches to diverse urban sensing task configurations.

Delving deeper, single-agent LLMs exhibit distinct trade-offs between feasibility and performance. Advanced models like GPT-4.1 and Claude-3.7 achieve high AIR in the \emph{Small} setting (e.g., $1.352\%$ and $1.165\%$ on T-Drive), but their SR declines sharply to $30.0\%$ and $45.0\%$ in the \emph{Large} setting due to frequent budget violations, suggesting an overly aggressive optimization strategy. In contrast, models such as GPT-3.5-turbo and DeepSeek-R1 adopt more conservative refinements, yielding higher SR but lower AIR. AgentSense, by comparison, achieves both high feasibility and strong improvement across all settings, for instance, attaining $100.0\%$ SR and $1.622\%$ AIR on Grab-Posisi \emph{Small}, and maintaining $95.0\%$ SR and $0.983\%$ AIR even on T-Drive \emph{Large}. These results highlight the benefits of AgentSense’s multi-agent iterative refinement, which balances performance gains with constraint adherence through cyclic verification, enabling robust and consistent enhancements beyond single-agent paradigms.

\vspace{-0.5em}
\subsection{Performance on Different Baselines (RQ2)}

\begin{table*}[!t]
\centering
\renewcommand{\arraystretch}{1.0}
\caption{Performance of AgentSense under different baseline planners across three task settings on T-Drive problem sets.}
\vspace{-1em}
\begin{tabular}{c|cccc|cccc|cccc}
\Xhline{1.0pt} 
\multirow{2}{*}{\textbf{\makecell{\\Baselines}}} 
& \multicolumn{4}{c|}{\textbf{Small}} 
& \multicolumn{4}{c|}{\textbf{Medium}} 
& \multicolumn{4}{c}{\textbf{Large}} \\ \cline{2-13}
& \makecell{Base \\ Obj. $\uparrow$} & \makecell{Final \\ Obj. $\uparrow$} & \makecell{AIR \\ (\%) $\uparrow$} & \makecell{Tokens \\ ($\times 10^5$) $\downarrow$} 
& \makecell{Base \\ Obj. $\uparrow$} & \makecell{Final \\ Obj. $\uparrow$} & \makecell{AIR \\ (\%) $\uparrow$} & \makecell{Tokens \\ ($\times 10^5$) $\downarrow$} 
& \makecell{Base \\ Obj. $\uparrow$} & \makecell{Final \\ Obj. $\uparrow$} & \makecell{AIR \\ (\%) $\uparrow$} & \makecell{Tokens \\ ($\times 10^5$) $\downarrow$}  \\
\Xhline{1.0pt}
RN                      & 4.505 & 4.630 & \textbf{2.775} & \textbf{1.620} & 4.706 & 4.784 & \textbf{1.657} & \textbf{2.770} & 5.029 & 5.221 & \textbf{3.818} & \textbf{5.733} \\
TVPG                    & 4.565 & 4.632 & 1.468 & 1.719 & 4.761 & 4.811 & 1.050 & 3.013 & 5.602 & 5.662 & 1.071 & 6.140 \\
TCPG                    & 4.588 & 4.628 & 0.872 & 1.968 & 5.118 & 5.178 & 1.172 & 3.207 & 5.544 & 5.617 & 1.317 & 6.505 \\
MSA                     & 4.662 & 4.674 & 0.257 & 2.356 & 5.086 & 5.136 & 0.983 & 4.873 & 5.631 & 5.677 & 0.817 & 6.953 \\
MSAGI                   & 4.702 & 4.713 & 0.234 & 2.524 & 5.090 & 5.101 & 0.212 & 4.860 & 5.682 & 5.699 & 0.299 & 7.834 \\
GraphDP~\cite{ji2016urban}  & \textbf{4.780} & \textbf{4.788} & 0.167 & 2.894 & \textbf{5.205} & \textbf{5.254} & 0.365 & 4.797 & \textbf{5.701} & \textbf{5.708} & 0.175 & 8.270 \\
\Xhline{1.0pt} 
\end{tabular}
\vspace{-0.5em}
\label{tab:diff_baseline}
\end{table*}

In this section, we investigate whether stronger baseline planners lead to better overall performance when combined with AgentSense. To this end, we evaluate AgentSense on 20 problem instances from the T-Drive dataset with six baseline algorithms: \textit{Random (RN)}, \textit{Task Value Priority Greedy (TVPG)}, \textit{Task Cost Priority Greedy (TCPG)}, \textit{Multi-Start Simulated Annealing (MSA)}, \textit{Multi-Start Simulated Annealing with Greedy Initialization (MSAGI)}, and an advanced \textit{Graph-based Dynamic Programming (GraphDP)} method. The first five are sourced from~\cite{wang2024urban}, while the GraphDP method is implemented accoring to~\cite{ji2016urban}. Detailed introduction are provided in Appendix~\ref{appendix:baseline_planner}.

Results in Table~\ref{tab:diff_baseline} shows that AgentSense consistently improves all baselines across all three settings. However, the degree of improvement is inversely related to baseline strength. For instance, AgentSense improves the \textit{RN} baseline by 2.775\% in the \emph{Small} setting and 3.818\% in the \emph{Large} setting, whereas gains on the more powerful baseline \textit{GraphDP} are minimal (0.167\% and 0.175\% respectively) and come with much higher token costs ($2.894 \times 10^5$ in \emph{Small} and $8.270 \times 10^5$ in \emph{Large}). This pattern reflects diminishing returns: as baseline quality approaches near-optimality, the space for refinement shrinks while the computational overhead of marginal improvements increases. Overall, weaker baselines such as \textit{TVPG} and \textit{TCPG} yield more cost-effective improvements with our framework, suggesting that it is often more efficient to start from a moderately strong baseline and leverage AgentSense for enhancement rather than beginning with an already highly optimized solution. In this paper, TVPG is adopted as the default planner.

\subsection{Performance on Disturbances (RQ3)}

\begin{table*}[!t]
\centering
\renewcommand{\arraystretch}{1.0}
\caption{Performance of AgentSense under different disturbances on 20 T-Drive problems. Metrics include \textit{SR} (Success Rate), \textit{AIR} (Average Improvement Rate), \textit{ANI} (Average Number of Iterations), and \textit{ACS} (Average Cost Savings).}
\vspace{-0.5em}
\begin{tabular}{l|cccc|cccc|cccc}
\Xhline{1.0pt} 
\multirow{2}{*}{\textbf{\makecell{\\Disturbances}}} 
& \multicolumn{4}{c|}{\textbf{Small}} 
& \multicolumn{4}{c|}{\textbf{Medium}} 
& \multicolumn{4}{c}{\textbf{Large}} \\ \cline{2-13}
& \makecell{SR \\ (\%) $\uparrow$} & ANI $\downarrow$ & \makecell{AIR \\ (\%) $\uparrow$} & ACS $\uparrow$
& \makecell{SR \\ (\%) $\uparrow$} & ANI $\downarrow$ & \makecell{AIR \\ (\%) $\uparrow$} & ACS $\uparrow$
& \makecell{SR \\ (\%) $\uparrow$} & ANI $\downarrow$ & \makecell{AIR \\ (\%) $\uparrow$} & ACS $\uparrow$ \\
\Xhline{1.0pt}
Budget change (increase)    & 100.0 & 5.6 & 5.855   & --    & 100.0 & 6.6 & 6.424   & --    & 100.0 & 8.5 & 3.227   & -- \\
Area blocked                & 100.0 & 7.4 & -0.551  & -0.10 & 100.0 & 7.2 & -0.121  & 0.10  & 95.0  & 8.3 & 0.000   & 0.00 \\
Priority area               & 100.0 & 6.1 & -0.882  & 0.00  & 100.0 & 7.9 & -0.320  & 0.20  & 100.0 & 8.9 & -0.182  & -0.10 \\
Mid-path visiting           & 100.0 & 7.3 & -0.667  & 0.10  & 95.0  & 8.5 & -0.783  & -0.26 & 90.0  & 9.3 & -1.260  & 3.00 \\
Worker unavailable          & 100.0 & 4.4 & -1.220  & 2.00  & 100.0 & 6.7 & -0.967  & 2.00  & 100.0 & 7.7 & -0.933  & -0.75 \\
New worker available        & 100.0 & 5.0 & 0.238   & -1.00 & 100.0 & 6.6 & 0.372   & -1.00 & 100.0 & 8.4 & 1.889   & 1.50 \\
Bad weather                 & 100.0 & 6.3 & -28.503 & 0.00  & 100.0 & 7.1 & -34.550 & 2.00  & 95.0  & 9.7 & -44.292 & 1.95 \\
\Xhline{1.0pt} 
\end{tabular}
\label{tab:disturbance}
\end{table*}

\begin{figure*}[!t]
        \centering
        \includegraphics[trim={5cm 4cm 3.5cm 3.5cm}, clip,width=0.95\linewidth]{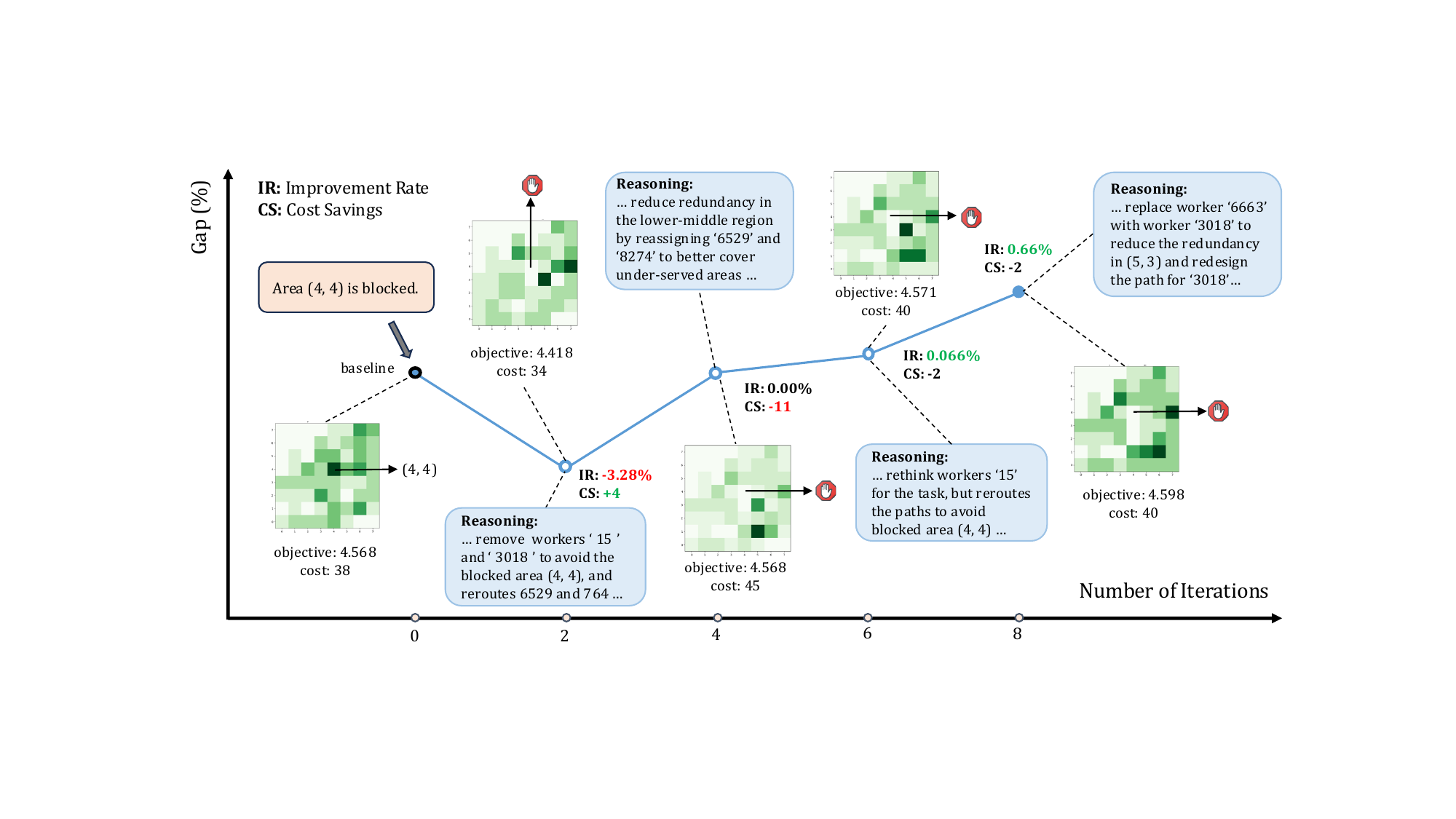}
        \vspace{-1.0em}
        \caption{Iterative improvement process for disturbance "Area blocked of (4, 4)".}
        \vspace{-0.5em}
\label{fig:iteration_result}
\end{figure*}

We evaluates AgentSense under seven types of dynamic disturbances commonly encountered in WPUS:
\begin{itemize}[leftmargin=*]
    \item \textbf{Budget change:} the budget dynamically increases / decreases.
    \item \textbf{Area blocked:} certain regions become temporarily inaccessible.
    \item \textbf{Priority area:} importance regions require priority coverage.  
    \item \textbf{Mid-path visit:} workers require additional mid-path visiting.
    \item \textbf{Worker unavailable:} one or more workers drop out.
    \item \textbf{New worker available:} additional workers join the task.
    \item \textbf{Bad weather:} adverse weather reduce worker mobility. 
\end{itemize}
While these seven cases are used for evaluation, AgentSense is not restricted to them. With suitable prompt design, the framework can flexibly accommodate other real-world disturbances.

 Table~\ref{tab:disturbance} shows that AgentSense demonstrates strong robustness across all disturbance types. In the \emph{Small} and \emph{Medium} settings, it consistently achieves perfect success rates (\textit{SR} = 100.0\%) and converges within relatively few iterations. The refined solutions generally remain feasible with negligible efficiency loss. For instance, under \textit{Area blocked}, the performance impact is minimal (\textit{AIR} = $-0.12\%$, \textit{ACS} = $0.10$ in \emph{Medium}), and both metrics stay close to zero in the \emph{Large} setting, indicating that AgentSense can reliably adapt baseline plans into high-quality, constraint-satisfying solutions. In the \emph{Large} setting, disturbance adaption becomes more challenging, reflected in slightly reduced success rates (e.g., \textit{SR} = 90.0\% for \textit{mid-path visiting} and 95.0\%). Nevertheless, the system still converges after several refinement rounds and maintains feasible solutions. Among all disturbances, \textit{bad weather} exerts the strongest negative influence on \textit{AIR} (e.g., $-28.503\%$ in \emph{Small}, $-44.292\%$ in \emph{Large}), since reduced worker mobility inherently limits achievable coverage. Even in this case, AgentSense preserves budget feasibility and leverages dynamic reallocation to mitigate performance degradation as much as possible.

\begin{figure*}[!t]
        \centering
        \includegraphics[trim={0cm 9.4cm 0cm 9.4cm}, clip,width=\linewidth]{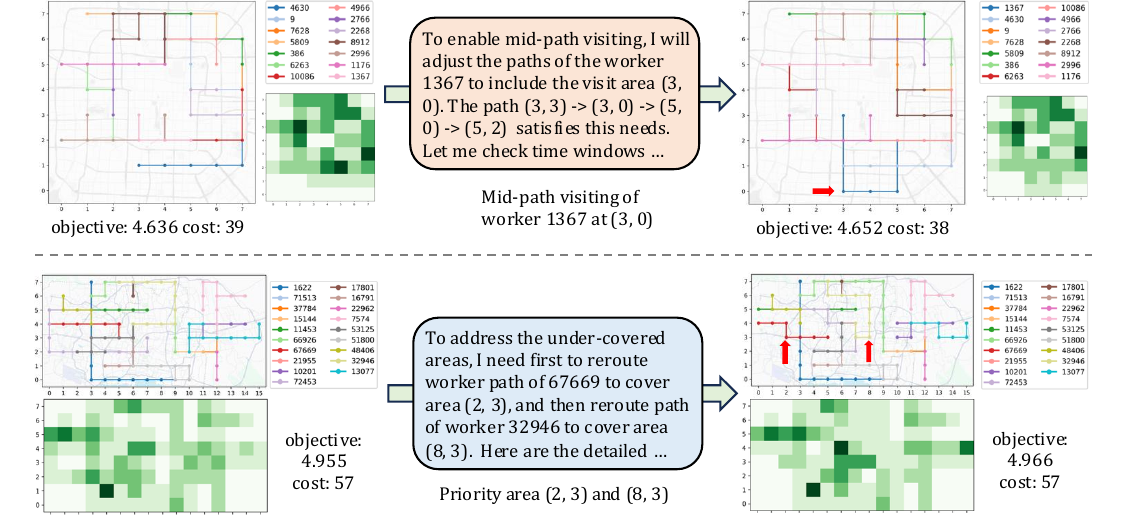}
        \caption{Adaptation results under different disturbances: (top) \emph{Mid-path visiting} in T-Drive \emph{Small} setting, and (bottom) \emph{Priority area} in Grab-Posisi \emph{Medium} setting. Significant route changes are highlighted with red arrows.}
        \vspace{-0.5em}
\label{fig:visual_example}
\end{figure*}

 To better illustrate the refinement process, Figure~\ref{fig:iteration_result} presents the optimization trajectory under an \textit{Area blocked} disturbance in the T-Drive \emph{Small} setting. Starting from an infeasible baseline, AgentSense incrementally reassigns workers and reroutes paths, with each step explained through natural language reasoning. The first adjustment avoided the blocked area but reduced data coverage, whereas later refinements improved coverage at the expense of sharply increased costs that exceeded the budget. After several corrections, the system converged to a feasible solution with higher coverage and budget compliance. This case highlights how iterative refinement enables AgentSense to recover from infeasibility and balance competing objectives under disturbances. Beyond this example, we further provide two additional results in Figure~\ref{fig:visual_example}, illustrating AgentSense’s adaptability to a \emph{Mid-path visiting} and a \emph{Priority area} disturbances. We highlight the significant changes in the routes by red arrows. Together, these cases demonstrate the framework’s robustness in handling diverse real-world disturbances.

\vspace{-0.5em}
\subsection{Ablation Studies (RQ4)}

\begin{figure}[!t]
\centering
\includegraphics[trim={0cm 0cm 0cm 0cm}, clip,width=0.98\linewidth]{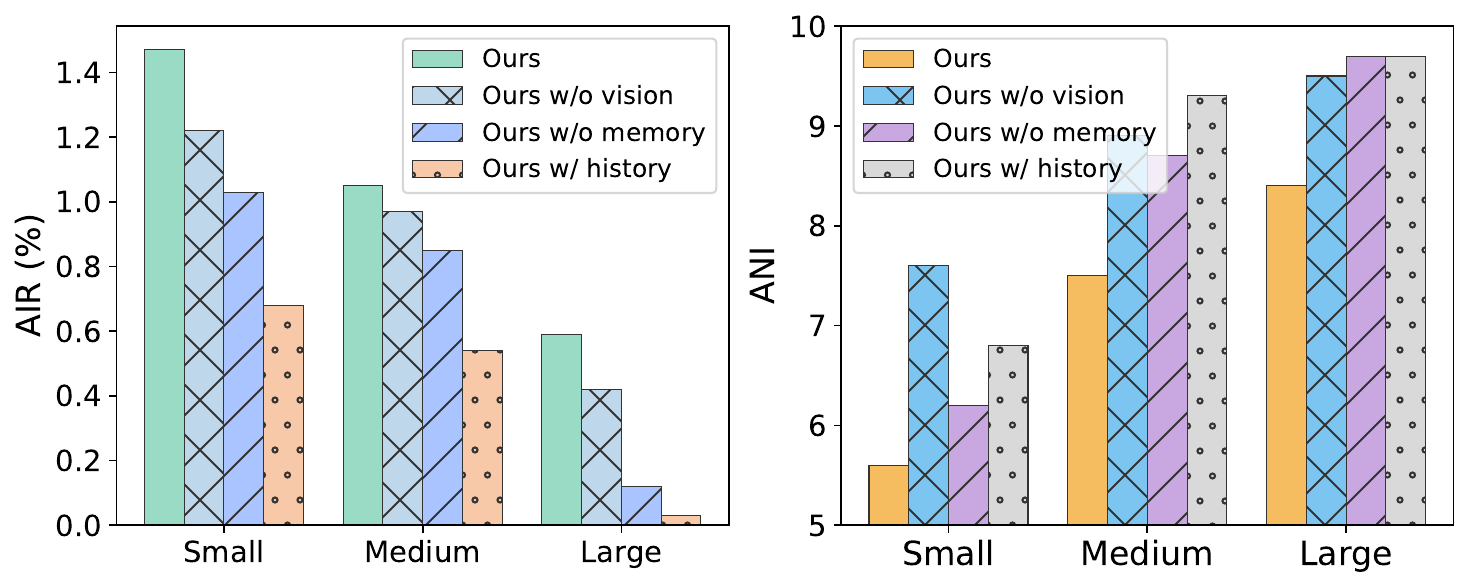}
\vspace{-0.5em}
\caption{Ablation studies of vision-enhanced evaluation and memory strategies.}
\vspace{-1.5em}
\label{fig:ablation}
\end{figure} 

In this section, we analyze how the vision module of the \emph{Eval Agent} and the structured memory maintained by the \emph{Memory Agent} affect the overall performance. Figure~\ref{fig:ablation} reports the ablation results, where the left panel shows the performance metric AIR and the right panel presents the convergence efficiency metric \emph{ANI}. The variant \emph{Ours w/ History} replaces the \emph{Memory Agent} with direct reuse of prior solutions as static references.

The results demonstrate that disabling either the vision module or the \emph{Memory Agent} leads to notable performance degradation, evidenced by reduced AIR and slower convergence (higher ANI). This underscores the critical role of these components, since vision module enables the \emph{Eval Agent} to reason over coverage distributions from a global perspective, facilitating faster identification of under-served regions and more informed refinements. Concurrently, the \emph{Memory Agent} leverages meta-operation-based memories to provide structured, transferable priors, effectively guiding the refinement process and minimizing redundant exploration. Additionally, we observe that directly reusing historical plans (\emph{Ours w/ History}) performs even worse than the no-memory baseline. This likely stems from the fact that raw solutions are structurally complex, making them hard for LLMs to interpret. Moreover, they consume substantial token budget, introducing noise that may impair decision-making. Overall, the complementary roles of \emph{Eval Agent} and \emph{Memory Agent} significantly enhance system performance: vision-based \emph{Eval Agent} improves evaluation accuracy and supports more meaningful suggestions for improvement, while \emph{Memory Agent} accelerates convergence and enhances robustness.

\vspace{-0.5em}
\subsection{Interpretability Analysis (RQ5)}

\begin{figure}[!t]
\centering
\includegraphics[trim={0cm 0cm 0cm 0cm}, clip,width=0.76\linewidth]{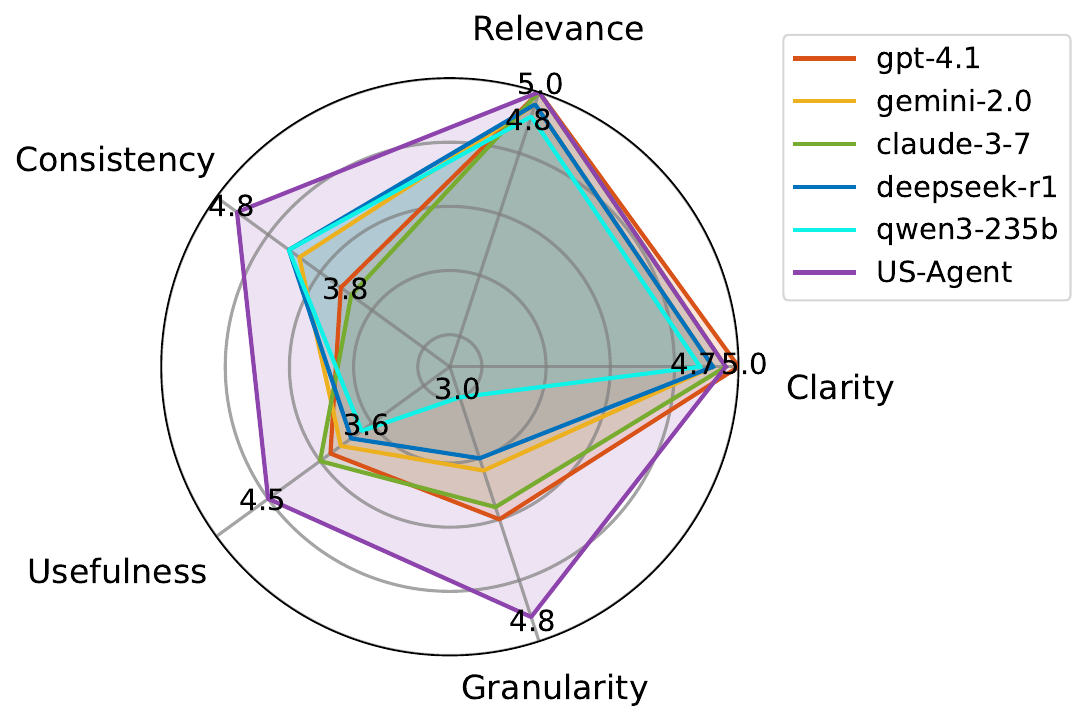}
\caption{Comparison of interpretability across five dimensions between AgentSense and baseline LLMs.}
\vspace{-1.5em}
\label{fig:radar_chart}
\end{figure} 

To systematically assess interpretability, we conducted a user study with 10 domain experts in urban sensing, who evaluated the explanations generated during the iterative refinement process. Explanations were rated from 0 to 5 following five dimensions: 
\begin{itemize}[leftmargin=*]
    \item \textbf{Clarity}: ease of understanding by humans.  
    \item \textbf{Relevance}: direct relation to the improved task solution.  
    \item \textbf{Consistency}: alignment with the actual planning behavior.  
    \item \textbf{Usefulness}: utility for supporting human decision-making.  
    \item \textbf{Granularity}: the level of refinement details.  
\end{itemize}
These criteria assess both linguistic quality and faithfulness to the agent’s reasoning and operations.

As shown in Figure~\ref{fig:radar_chart}, AgentSense obtains the highest scores across nearly all five dimensions, consistently surpassing single-agent LLM baselines. Expert evaluations revealed that while GPT-4.1 and Claude-3.7 produced fluent explanations, they occasionally neglected feasibility constraints, leading to lower \emph{Consistency}. Gemini-2.0 favored brevity but often lacked sufficient elaboration, resulting in reduced \emph{Granularity}. Open-source models such as DeepSeek-R1 and Qwen3-235B generated explanations with moderate alignment but limited actionable depth. In contrast, AgentSense’s iterative refinement mechanism—linking \textit{Solver Agent} updates with systematic feedback from the \emph{Eval Agent}—produced explanations judged to be both faithful to the underlying operations and sufficiently detailed to support expert reasoning. These findings demonstrate that the multi-agent design enhances not only solution quality but also transparency, yielding explanations that domain experts considered more interpretable, trustworthy, and practically valuable for urban sensing applications.

\vspace{-1em}
\section{Conclusion and Future Work}
In this work, we presented \textbf{AgentSense}, a hybrid, training-free framework for web-based participatory urban sensing. By integrating a classical planner with an LLM-driven multi-agent refinement module, AgentSense addresses two central challenges in WPUS: context-aware generalization across heterogeneous and dynamic conditions, and transparent interpretability that elucidates the reasoning behind solution generation. Extensive experiments on two large-scale mobility datasets under varied disturbance settings demonstrate that AgentSense consistently outperforms single-agent LLMs and traditional approaches in both accuracy and adaptability. Expert evaluations further confirm the clarity and reliability of its explanations. Looking forward, we outline two key directions for future work. First, we plan to incorporate geographical attributes of sensing grids to enable personalized path planning, thereby improving efficiency and contextual awareness in task allocation. Second, we aim to enhance scalability by adapting the framework to smaller language models (e.g., 7B or 3B parameters), making AgentSense more resource-efficient and practical for real-world urban sensing deployments.



\bibliographystyle{ACM-Reference-Format}
\bibliography{ref}

\appendix

\clearpage
\twocolumn[{
	\renewcommand\twocolumn[1][]{#1}
	\begin{center}
 \textbf{\fontsize{15}{48}\selectfont Appendix}
  \end{center}
        \vspace{0.5cm}
}]

\section{Baseline Planner}
\label{appendix:baseline_planner}

This appendix summarizes the six baseline planners adopted in our study. All methods produce an initial feasible schedule $\mathcal{S}_0$ that meets budget and temporal constraints before refinement. The first five are reproduced from~\cite{wang2024urban}, and the \emph{GraphDP} algorithm follows~\cite{ji2016urban}.

\begin{itemize}[leftmargin=*]
\item \textbf{Random (RN)}.
For each worker, an initial route is generated from its origin to a fixed destination using the \textit{Nearest Neighbor} heuristic. Sensing tasks are then randomly inserted into feasible positions along the route until the worker’s budget is exhausted. This provides a purely stochastic baseline that satisfies route and cost constraints.

\item \textbf{Task Value Priority Greedy (TVPG)}.
TVPG starts from the same \textit{Nearest Neighbor} route construction but greedily inserts tasks that yield the highest marginal coverage gain when added to any feasible position of a worker’s route. When multiple tasks offer equal gain, the task with the lower incentive cost is selected.

\item \textbf{Task Cost Priority Greedy (TCPG)}.
Similar to TVPG, TCPG also evaluates feasible insertions but prioritizes tasks with lower incentive costs. If several tasks share the same cost, the one that provides higher coverage gain is preferred.

\item \textbf{Multi-Start Simulated Annealing (MSA)}.
MSA performs iterative local search over complete worker routes through random operations such as swapping, insertion, and reversal. Each candidate modification must preserve route feasibility (budget and connectivity). Multiple random restarts are executed to escape local minima and improve robustness.

\item \textbf{Multi-Start Simulated Annealing with Greedy Initialization (MSAGI)}.
MSAGI enhances MSA by initializing each run with the greedy solution from TVPG instead of random routes. This improves convergence speed and solution quality by combining heuristic initialization with metaheuristic search.

\item \textbf{Graph-based Dynamic Programming (GraphDP)}.
GraphDP constructs a time-layered location graph for each worker and applies dynamic programming to search near-optimal paths that maximize spatial–temporal coverage based on marginal entropy gains. Then worker replacement further improves overall coverage by swapping low-contribution workers with better candidates until no significant gain is achieved.
\end{itemize}

\section{System Prompts}

This section presents the complete prompt designs used for the three core agents in the \emph{AgentSense} framework. Each prompt defines the agent’s role and goal, input structure, and the output format. Specifically, Section~\ref{appendix:solver_agent_prompt} details the \emph{Solver Agent} prompt that guides solution generation and modification; Section~\ref{appendix:eval_agent_prompt} describes the \emph{Eval Agent} prompt for assessing solution quality and constraint compliance; and Section~\ref{appendix:memory_agent_prompt} provides the \emph{Memory Agent} prompt used to retrieve and reuse prior meta-operations for experience-aware refinement.

\subsection{Solver Agent Prompt}
\label{appendix:solver_agent_prompt}

\begin{tcolorbox}[
    enhanced,
    colback=gray!10!white, 
    colframe=gray!80!white, 
    colbacktitle=gray!70!white, 
    coltitle=white, 
    arc=2mm, 
    boxrule=1pt, 
    left=1mm,   
    right=1mm,  
    top=1mm,    
    bottom=1mm, 
]
\small 

\textbf{\# Role and Goal:} \\
You are an expert in solving the urban sensing task assignment problem. Your goal is to improve a baseline solution according to the given instructions while ensuring compliance with budget and feasibility constraints.


\textbf{\# Inputs:} \\
You will be given the following information for your task: 

- \textbf{Task settings}: global information of the task, such as grid size, total budget and task time range.

- \textbf{Worker info}: information of the worker candidates, such as origin, destination, time windows, and speed.

- \textbf{Baseline solution}: task assignments from baseline planner.

- \textbf{Instructions}: "continue optimize" or one of [disturbance types].


\textbf{\# Constraints:}

You need to consider the following constraints:

\quad - Worker must start/end at origin/dest within time window.
    
\quad - Worker move in four direction: up, down, left, right, or stay.
    
\quad - Total cost should be within the total budget.

\textbf{\# Tools you can use:}

\quad - \textit{validate\_worker\_paths}

\quad \quad parameters: \{"worker\_paths": \{...\}, "worker\_infos": [...], ...\}.

\quad \quad return: \{"overall\_feasible": bool, "validation\_results": ...\}.


\textbf{\# Final output:} 

Organize your final answer as following:

\quad - \textbf{think\_process}: up to 200 words summary of reasoning process. 

\quad - \textbf{refined\_solution}: \{"worker\_id": [[x, y, t], [x, y, t], ...], ...\} .

\end{tcolorbox}

\subsection{Eval Agent prompt}
\label{appendix:eval_agent_prompt}

\begin{tcolorbox}[
    enhanced,
    colback=gray!10!white, 
    colframe=gray!80!white, 
    colbacktitle=gray!70!white, 
    coltitle=white, 
    arc=2mm, 
    boxrule=1pt, 
    left=1mm,   
    right=1mm,  
    top=1mm,    
    bottom=1mm, 
]
\small 

\textbf{\# Role and Goal:} \\
You are an expert in urban sensing task allocation. Given a candidate solution, your task is to evaluate its performance using quantitative metrics and visual evidence, and then provide clear, specific, and actionable recommendations for improvement.


\textbf{\# Inputs:} \\
The baseline and refined solutions in the following format:

\quad - \{"worker\_id": [[x, y, t], [x, y, t], ...], ...\}. A dictionary where each key is a worker ID, and the value is a list of grid positions \([x, y, t]\) representing the worker’s assigned trajectory over space and time.


\textbf{\# Instruction:} 

\quad - Use tools to check how well the solution handles disturbances. 

\quad - Use tools to compute the objective value and cost of the solution. 

\quad - Use tools to generate a visualization that compares the refined solution with the baseline. 

\quad - When providing suggestions, base them on both the quantitative evaluation and the visual evidence.

\textbf{\# Tools you can use:}

\quad -\textit{check\_disturbance\_handling(...)}

\quad - \textit{compute\_objective(...)}

\quad - \textit{compute\_cost(...)}

\quad - \textit{visual\_analysis(...)}


\textbf{\# Final output:} 

Organize your final answer as following:

\quad - \textbf{eval\_summary}: a concise summary of the evaluation results.

\quad - \textbf{advice}: specific, actionable advice for improving the solution.

\end{tcolorbox}

\subsection{Memory Agent prompt}
\label{appendix:memory_agent_prompt}

\begin{tcolorbox}[
    enhanced,
    colback=gray!10!white, 
    colframe=gray!80!white, 
    colbacktitle=gray!70!white, 
    coltitle=white, 
    arc=2mm, 
    boxrule=1pt, 
    left=1mm,   
    right=1mm,  
    top=1mm,    
    bottom=1mm, 
]
\small 

\textbf{\# Role and Goal:} \\
You are a memory manager in urban sensing task allocation. Your goal is to detect differences between baseline and refined solutions, extract the most critical adaptation as a \textit{meta-operation}, and store it in memory so that the system can leverage this knowledge to improve future urban sensing solutions.


\textbf{\# Inputs:} \\
You will be provided with the baseline and refined solutions along with their evaluation metrics in the following format:

\quad - \textbf{Solution}: \{"worker\_id": [[x, y, t], [x, y, t], ...], ...\}. A dictionary where each key is a worker ID, and the value is a list of grid positions \([x, y, t]\) representing the worker’s assigned trajectory over space and time.

\quad - \textbf{Metrics}: \textit{covered\_count}, \textit{entropy}, \textit{objective\_value}, \textit{cost}.


\textbf{\# Instruction:}

\quad - Operation types: \textit{add\_worker}, \textit{remove\_worker}, \textit{modify\_path}, \textit{other}. 

\quad - First, detect all differences between baseline and refined solutions. 

\quad - Next, determine the most significant difference.

\quad - Finally, combine this difference with the changes observed in the metrics to produce the output.


\textbf{\# Final output:} 
Organize your final answer as following:

\quad - \textbf{operation\_type}: the category of the difference between solutions.

\quad - \textbf{operation\_details}: a concise description of what changed and how it affected the metrics.

\end{tcolorbox}

\section{More Experimental Results for Temperature}

\begin{figure}[!t]
\centering
\includegraphics[trim={0cm 0cm 0cm 0cm}, clip,width=\linewidth]{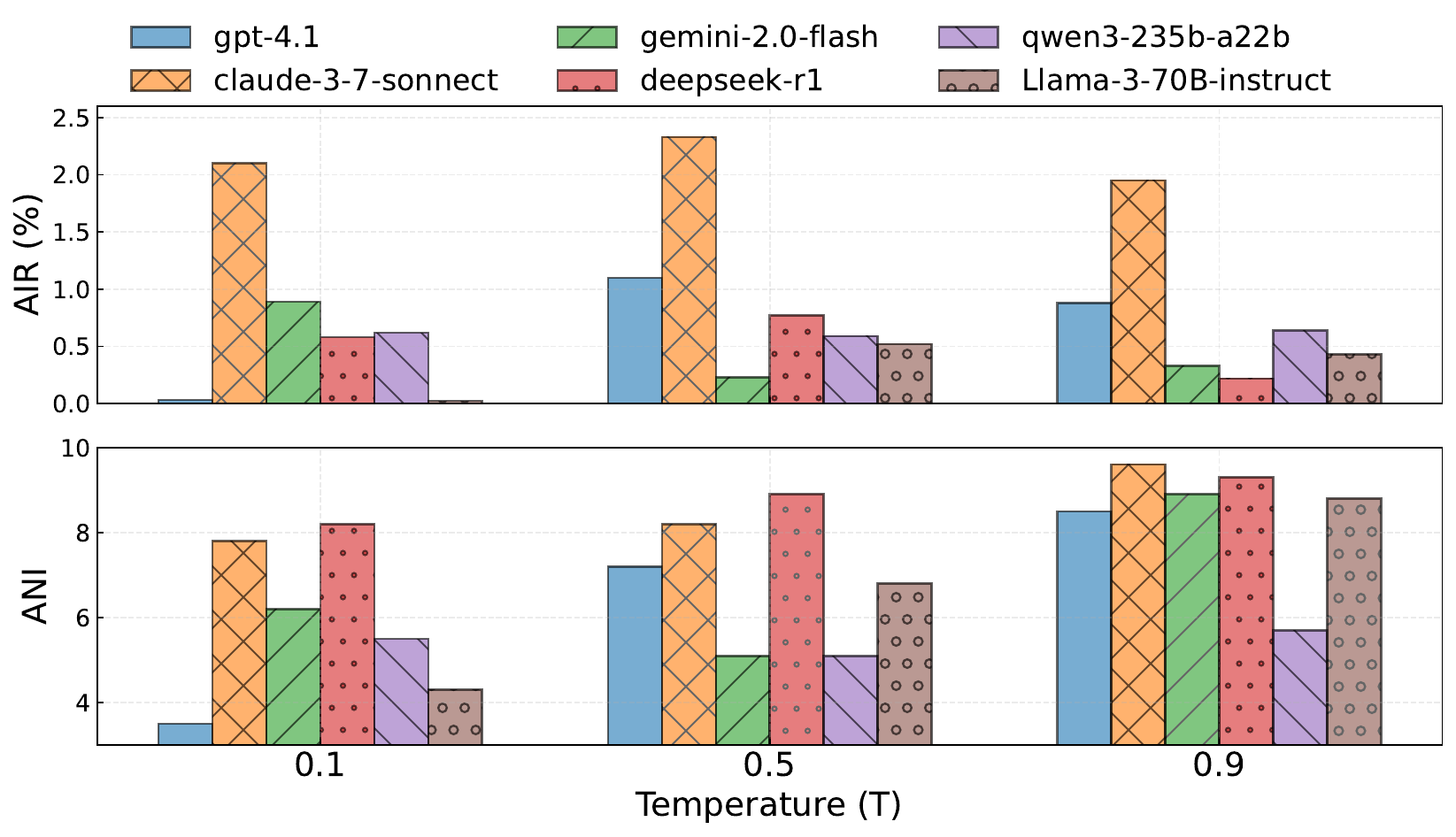}
\vspace{-1.0em}
\caption{Comparison of different base LLMs within AgentSense under varying temperature settings in the T-Drive \emph{Medium} configuration.}
\vspace{-0.5em}
\label{fig:diff_temp}
\end{figure}

We also evaluate AgentSense with different base LLMs and analyze the impact of decoding temperature. As shown in Figure~\ref{fig:diff_temp}, the optimal temperature varies across models. GPT-4.1 achieves its highest AIR ($1.10\%$) at $T=0.5$, while Gemini-2.0-Flash performs best at $T=0.1$, suggesting that more deterministic decoding improves stability. Claude-3-7-Sonnet benefits from moderate exploration, reaching $2.33\%$ AIR at $T=0.5$, but its performance declines at $T=0.9$, indicating that excessive randomness reduces robustness. In contrast, open-source models such as Qwen3 and Llama-3 remain relatively stable across all temperatures. Moreover, the ANI metric reveals that lower temperatures generally lead to faster convergence, as fewer iterations are required to reach valid improvements. Overall, these results highlight two key insights: (i) AgentSense is compatible with both proprietary and open-source LLMs, consistently enforcing budget feasibility across all settings, and (ii) careful tuning of decoding temperature is essential to balance exploration and exploitation, as different LLMs exhibit distinct sensitivities to sampling variability.

\section{More Results for Disturbance Adaptation}

\begin{figure*}
\centering
\includegraphics[trim={0.5cm 4.5cm 0.5cm 0cm}, clip,width=0.92\linewidth]{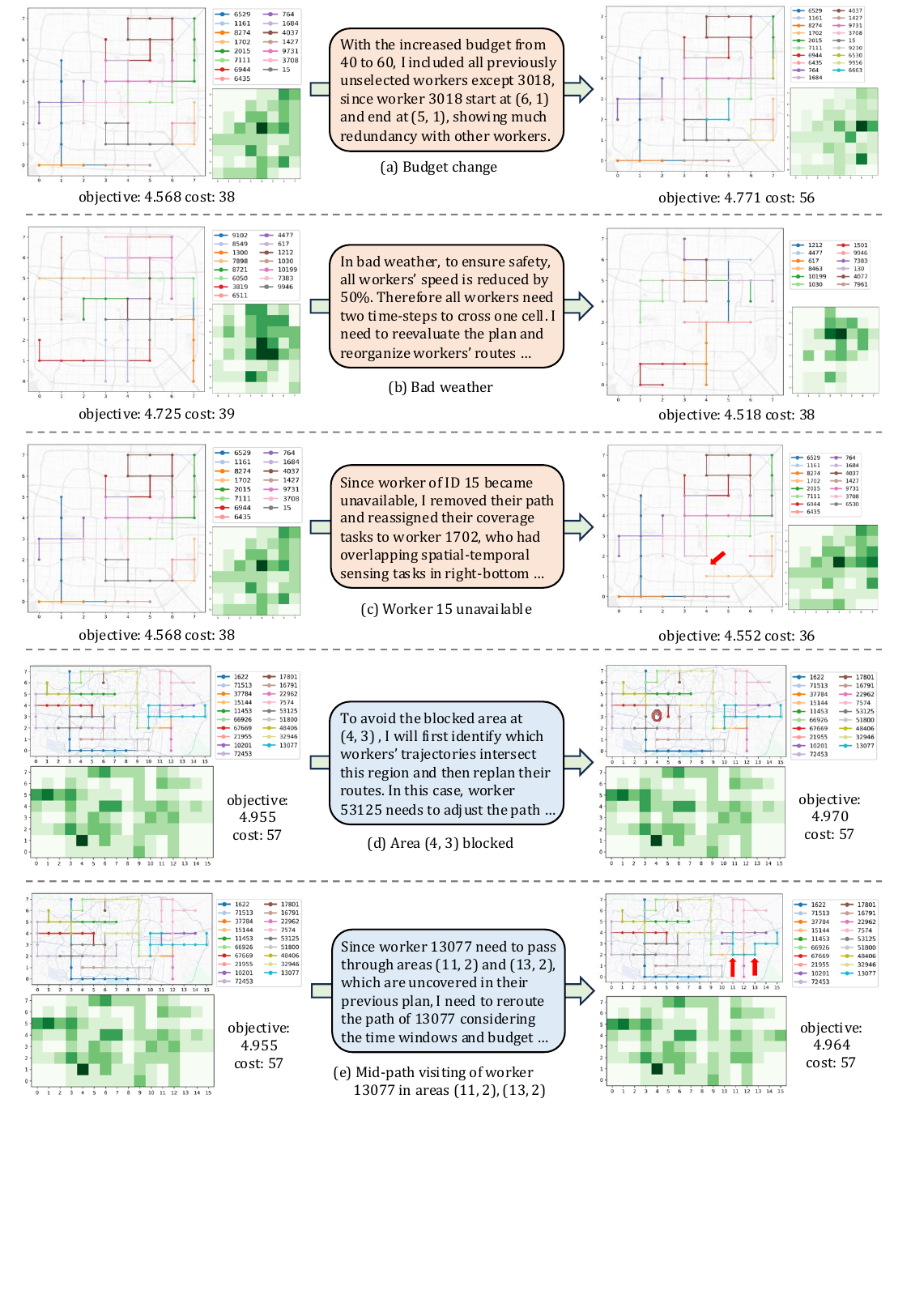}
\caption{Visualization and reasoning results of AgentSense under different types of disturbances. The first three cases are from the T-Drive \emph{Small} setting, and the last two are from the Grab-Posisi \emph{Medium} setting.}
\label{fig:visual_results}
\end{figure*}

Figure~\ref{fig:visual_results} presents five representative types of disturbances considered in this study: \emph{budget change}, \emph{bad weather}, \emph{worker unavailable}, \emph{area blocked}, and \emph{mid-path visiting}. These cases illustrate how AgentSense adapts to diverse real-world disturbances through interpretable reasoning and adaptive replanning. In the T-Drive \emph{Small} setting, AgentSense responds to budget rising by adding workers while avoiding redundant routes, reorganizes trajectories under bad weather to account for slower movement, and reassigns coverage when a worker becomes unavailable. In the Grab-Posisi \emph{Medium} setting, it reroutes affected paths to bypass a blocked area and adjusts trajectories to fulfill new mid-path visiting requests without violating constraints. Across all scenarios, AgentSense maintains feasible, high-coverage solutions and provides clear reasoning for each refinement step, highlighting its robustness and interpretability under dynamic urban conditions.

\end{document}